\documentclass[sigconf]{acmart}

\usepackage{hyperref}       
\usepackage{url}            
\usepackage{amsfonts}       

\usepackage{comment}

\usepackage{multirow}
\usepackage[font=small,labelfont=bf,tableposition=top]{caption}

\usepackage{blindtext}

\usepackage{epsfig}
\usepackage{graphicx}

\usepackage{amsmath}
\usepackage{amssymb}

\usepackage{graphics}

\newcommand{\systemname}{{\it $Fed^2$}\xspace}

\graphicspath{{_fig/}}

\title{Fed$^2$: Feature-Aligned Federated Learning}

\author{Fuxun Yu}
\email{fyu2@gmu.edu}
\affiliation{%
\institution{George Mason University}
\city{Fairfax}
\state{VA}
\postcode{22030}
\country{USA}}

\author{Weishan Zhang}
\email{wzhang23@gmu.edu}
\affiliation{%
\institution{George Mason University}
\city{Fairfax}
\state{VA}
\postcode{22030}
\country{USA}}

\author{Zhuwei Qin}
\email{zqin@gmu.edu}
\affiliation{%
\institution{George Mason University}
\city{Fairfax}
\state{VA}
\postcode{22030}
\country{USA}}

\author{Zirui Xu}
\email{zxu21@gmu.edu}
\affiliation{%
\institution{George Mason University}
\city{Fairfax}
\state{VA}
\postcode{22030}
\country{USA}}

\author{Di Wang}
\email{wangdi@microsoft.com}
\affiliation{%
\institution{Microsoft}
\city{Seattle}
\state{WA}
\postcode{98052}
\country{USA}}

\author{Chenchen Liu}
\email{ccliu@umbc.edu}
\affiliation{%
\institution{University of Maryland, Baltimore County}
\city{Baltimore}
\state{MD}
\postcode{21250}
\country{USA}}

\author{Zhi Tian}
\email{ztian1@gmu.edu}
\affiliation{%
\institution{George Mason University}
\city{Fairfax}
\state{VA}
\postcode{22030}
\country{USA}}

\author{Xiang Chen}
\email{xchen26@gmu.edu}
\affiliation{%
\institution{George Mason University}
\city{Fairfax}
\state{VA}
\postcode{22030}
\country{USA}}

\begin{abstract}

Federated learning learns from scattered data by fusing collaborative models from local nodes.
	However, conventional coordinate-based model averaging by FedAvg ignored the random information encoded per parameter and may suffer from structural feature misalignment.
In this work, we propose \systemname, a feature-aligned federated learning framework to resolve this issue by establishing a firm structure-feature alignment across the collaborative models.
	\systemname is composed of two major designs:
	First, we design a feature-oriented model structure adaptation method to ensure explicit feature allocation in different neural network structures.
	Applying the structure adaptation to collaborative models, matchable structures with similar feature information can be initialized at the very early training stage.
	During the federated learning process, we then propose a feature paired averaging scheme to guarantee aligned feature distribution and maintain no feature fusion conflicts under either IID or non-IID scenarios.
	Eventually, \systemname could effectively enhance the federated learning convergence performance under extensive homo- and heterogeneous settings, providing excellent convergence speed, accuracy, and computation/communication efficiency.

\end{abstract}



\copyrightyear{2021}
\acmYear{2021}
\setcopyright{acmcopyright}\acmConference[KDD '21]{Proceedings of the 27th ACM
SIGKDD Conference on Knowledge Discovery and Data Mining}{August 14--18,
2021}{Virtual Event, Singapore}
\acmBooktitle{Proceedings of the 27th ACM SIGKDD Conference on Knowledge
Discovery and Data Mining (KDD '21), August 14--18, 2021, Virtual Event, Singapore}
\acmPrice{15.00}
\acmDOI{10.1145/3447548.3467309}
\acmISBN{978-1-4503-8332-5/21/08}

\begin{document}
\fancyhead{}

\maketitle

\section{Introduction}
\label{sec:intro}

\begin{figure*}[t]
	\begin{center}
	\includegraphics[width=6.5in]{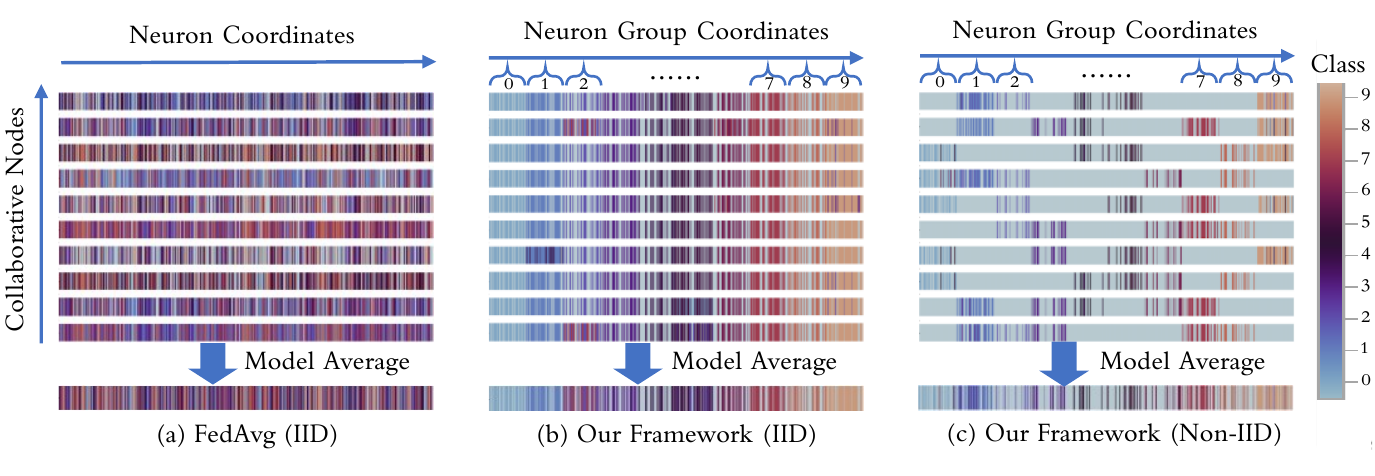}
	\vspace{-4mm}
	\caption{Feature encoding visualization of one sampled convolutional layer across ten clients during FL.
		The color of each neuron is determined by its top response class to indicate its learned feature. 
		\textbf{(a)} The original FedAvg with chaotic feature encoding can suffer from feature averaging conflicts among different nodes. 
		\textbf{(b) \& (c)} In contrast, our framework enforces structurally-aligned feature encoding by adopting group convolution and alleviates the averaging conflicts in both IID and non-IID cases.
		Experiments are conducted with ten collaborative nodes (VGG9 on CIFAR10). IID: Each node has local data of 10 classes. Non-IID: Each node has local data of only 5 classes.
		}
	\label{fig:motivation}
	\vspace{-3mm}
	\end{center}
\end{figure*}

Federated Learning (FL) has achieved great popularity among various distributed deep learning frameworks due to its superior collaboration flexibility, communication efficiency, and performance robustness in vision and language learning scenarios~\cite{fl, fl1, fl2, fl3}.
	It is commonly achieved by multiple FL nodes' collaboration through Federated Averaging (FedAvg), which generates a global model by periodically averaging local models' parameters.
	Specifically, FedAvg follows a coordinate-based weight averaging manner~\cite{fl, fedma}.
	Different local models' weights in the same layer and same index (i.e., coordinates) are averaged to be the global model's weight.

Although widely adopted, FedAvg still suffers from accuracy drop due to a common issue called weight divergence~\cite{non-iid0, pfnm, iclr}. 
Especially in non-IID scenarios, highly skewed data distribution across nodes can cause distinct weight values at the same coordinates, thus hurting the model averaging performance.
Recent works elaborate one potential reason of such weight divergence is the DNN ``permutation invariance'' property.
	Specifically, given a DNN model, the set of parameters in its convolutional and fully-connected layers could be arbitrarily permuted with different permutation sequences, while still yielding the same computational results~\cite{fedma}.
Due to the permutation invariance property, weight matrices of different local FL models may not be fully-aligned by coordinates. 
	Thus, coordinate-based FedAvg will incur weight averaging conflicts and lead to sub-optimal FL accuracy, which is commonly observed as the weight divergence issue.

Many optimization methods are proposed to alleviate the weight divergence issue by parameter-oriented weight matching, such as representation matching~\cite{rmatching}, Bayesian matching~\cite{bayesian}, FedMA~\cite{fedma}, {etc}.
	Although these works have different designs, such as weight aligning by minimizing MSE distance~\cite{fedma}, or activation aligning~\cite{rmatching}, they share the similar methodology: After each local model training epoch, they first evaluate the parameter similarity across local models, and then re-permute the weight matrices so that approximate weights could be averaged together.

Although outperforming native FedAvg, these methods still have certain limitations, such as inaccurate parameter similarity, extra computation/communication overhead and compromised data privacy, etc. 
	Specifically, current methods’ matching accurateness highly depends on the selected similarity metric and the operation targets. For example, [11, 18] use MSE loss with Euclidean distance in weight or activation matrices. 
	But two weight matrices with a small distance may not necessarily mean they carry the same information and feature.
	Therefore, common parameter matching methods can still suffer from the feature-level misalignment.

To tackle these limitations, we propose {\systemname}, a feature-aligned federated learning framework.
Fig.~\ref{fig:motivation} demonstrates a set of feature visualization and illustrates the different feature alignment effect between FedAvg and our proposed \systemname framework.
	As shown in Fig.~\ref{fig:motivation} (a), FedAvg’s local models suffer from significant feature-level mismatching. 
	The coordinate-based FedAvg in such case will thus incur dramatic feature conflicts and cause convergence performance degradation.
	By contrast, with the proposed \systemname, our models' learned feature distribution conform to strict structural alignment without any averaging conflicts.
	Even in extreme non-IID scenarios, \systemname still maintains consistent feature alignment among local models, thus providing superior federated learning performance than prior works including higher convergence rate, accuracy, etc.

Specifically, we make the following contributions:
\begin{itemize}
	\item First, we promote the previous weight-level matching methods to a feature-level alignment method by defining an feature interpretation method. Such a method analyzes and qualitatively shows the feature-level misalignment issue in current coordinate-based FedAvg algorithm;

	\vspace{1mm}
	\item We then propose a controllable feature allocation methodology by combining feature isolation and gradient redirection techniques. Such controllable feature allocation is achieved by a group-wise convolution and fully-connected structure adaptation, which can pre-align the feature and model structure even before the training process;

	\vspace{1mm}
	\item Eventually, we design a feature-aligned FL framework --- \systemname, which is composed of feature-oriented structure adaptation and model fusion algorithm. By maintaining consistent feature alignment throughout FL training, \systemname could achieve superior performance than FedAvg and other weight-matching methods.
	
\end{itemize}

We conduct extensive experiments on general datasets (CIFAR10, 100) with varied architectures (VGG9, VGG16 and MobileNet).
	The experimental results of \systemname demonstrate significant improvement in both convergence speed and accuracy, outperforming previous state-of-the-art works by large margins (+2.5\%$\sim$4.6\%) while having smaller computation and communication cost.
	Even under highly-skewed non-IID scenarios, our work still performs effective and robust feature alignment and ensures the near-optimal FL convergence accuracy when most previous methods fail to do so.

\section{Background and Related Work}
\label{sec:2}

\vspace{1mm}
\subsection{Federated Learning with FedAvg}
Conventional FL frameworks usually adopt the Federated Averaging algorithm (FedAvg)~\cite{fl} to collect distributed weight parameters from local nodes and fuse for the global DNN model:
\begin{equation}
	\Omega^{e+1}_{(l, i)} = \sum\nolimits_{n=1}^{N} \frac{1}{N}~\omega_{(n,l,i)}^{e},
	\label{eq:fedavg}
\end{equation}
where $\Omega$ and $\omega_{i}$ are the global weights and local weights from the $n^{th}$ node ($N$ nodes in total), $l$ and $i$ denote the layer and weight indexing for parameter coordination.
	$e$ denotes the epoch-wise weight averaging cycle, which FL leverages to facilitate the communication efficiency compared to iteration-wise averaging.

The FedAvg formulation in Eq.~\ref{eq:fedavg} implicitly defines a \textbf{\textit{coordinate-based parameter averaging}} for distributed local DNN models, \textit{i.e.}, weights of the same coordinates $(l, i)$ in these models are strictly designated to be averaged across the collaborative training process.

\subsection{Neural Network Permutation Invariance}
However, recent research works have proved that parameter coordination is an inaccurate DNN model fusion guidance by revealing a particular DNN property -- \textbf{\textit{weight permutation invariance}}~\cite{bayesian, pfnm, fedma}.
	Such a property shows that the DNN weight matrix can be structurally shuffled while maintaining lossless calculation results.

Specifically, a weight matrix $\omega$ can be decomposed as $\omega\Pi$, where $\omega$ indicates the parameter value only and $\Pi$ defines the coordinate permutation matrix.
	When $\Pi = \textbf{1}$ (identity matrix), no coordinate permutation is applied to the weight matrix, \textit{i.e.}, $\omega\textbf{1} = \omega$.
	While $\textbf{1}$ can be further decomposed a pair of permutation matrices ($\Pi~\Pi^T$)\footnote{We could construct any permuted identity matrix $\Pi$ by shuffling the \textbf{1} elements to be non-diagonal but maintains the full rank.}, the lossless permutation can be formulated as:
\begin{equation}
	\omega~\textbf{1} = \omega~\Pi~\Pi^T = \omega.
	\label{eq:invar}
\end{equation}
Without loss of generality, considering a DNN model composed of two consecutive layers (layer weights as $\omega_{l}$ and $\omega_{l+1}$) as an example, the output $F(X)$ could be formulated as\footnote{We use fully connected layers as an example. Convolutional layers could also be transformed to similar matrix multiplication calculation by \textit{im2col}.}:
\begin{equation}
	F(X) = \omega_{l+1}~(\omega_{l}~X).
\end{equation}
	According to Eq.~\ref{eq:invar}, applying any permutation matrix together with its transpose onto $\omega_{l+1}$ incurs no output influence:
\begin{equation}
	F(X) = (\omega_{l+1}~\Pi_{l+1}~\Pi_{l+1}^T)~(\omega_{l}~X) = (\omega_{l+1}~\Pi_{l+1})~(\Pi_{l+1}^T~\omega_{l})~X.
\end{equation}
Therefore, the original DNN weights of two layers $\omega_{l+1}, \omega_{l}$ could be losslessly re-permuted as $(\omega_{l+1}~\Pi_{l+1})$ and $(\Pi_{l+1}^T~\omega_{l})$, resulting in individual weight $\omega_i$'s allocation variation in a layer.

\subsection{FedAvg \textit{vs.} Permutation Invariance}
The permutation invariance property implies that a weight parameter could be permuted with arbitrary coordinates in a layer, which conflicts with the coordinate-based parameter averaging in FedAvg.

Specifically, suppose the DNN models of two consecutive layers ($l, l+1$) from $N$ local nodes all learn the same function $F(\cdot)$:
\begin{equation}
	\begin{split}
		F(\cdot) = & (\omega_{(0,l+1)} \Pi_0)~(\Pi_0^T \omega_{(0,l)}) = \cdots = (\omega_{(n,l+1)} \Pi_n)~(\Pi_n^T \omega_{(n,l)})\\
		= & \cdots = (\omega_{(N,l+1)} \Pi_N)~(\Pi_N^T \omega_{(N,l)}),~\hspace{4mm}~n \in (0, N).
		\label{eq:1}
	\end{split}
\end{equation}
Even we assume $N$ models can have the same weight values composition (\textit{i.e.}, $\omega_{(n,l)}$ = $\omega_l$), their coordinate matrices $\Pi_n$ could be highly different as these models are trained separately during the local training epoch:
\begin{equation}
	\Pi_0 \neq~\dots~\neq \Pi_n \neq~\dots~\neq \Pi_N,~n \in (0, N).
\end{equation}
Therefore, the weight parameter learning a particular information may have diverse in-layer coordinate $i$ across different local models.

As FedAvg still conducts rigid coordinate-based averaging:
\begin{equation}
	\Omega_{(l+1, i)} = \sum\nolimits_{n=1}^{N} \frac{1}{N}~\omega_{(n,l,i)}\Pi_n;
	\Omega_{(l, i)} = \sum\nolimits_{n=1}^{N} \frac{1}{N}~\Pi_n^T\omega_{(n,l,i)},
\end{equation}
the averaged weights can hardly match with each other, nor the corresponding information across $N$ models.

The permutation invariance gives a new explanation perspective to commonly-known FL issues, such as weight divergence and accuracy degradation, especially in non-IID data distributions where local models are even learning non-uniform information~\cite{non-iid0,fedprox,iclr}.

\subsection{Weight-Level Alignment (WLA)}
The permutation invariance not only explains the FedAvg issues, but also serves as a DNN model configuration tool to motivate many FL optimization works~\cite{fedma, bayesian, pfnm, rmatching}.
	These works identify the parameters with corresponding information across local DNN models with certain similarity metrics (\textit{e.g.}, \textit{MSE}) and leverage a lossless permutation matrix to \textbf{\textit{structurally align the parameters' allocation for matched information fusion}}.

Taking the two-layer DNN model as an example, this process can be formulated as re-permuting the $l^{th}$ layer weight matrix $\omega_{(n,l)}$ on the $n^{th}$ node with a re-permutation matrix $\Pi_{trans}$ to minimize the selected distance metric --- $D$ with the global weight $\Omega_l$:
\begin{equation}
	\begin{split}
		\omega_{(n, l+1)}^{aligned} = & \omega_{n, l+1}\Pi_{trans}, s.t.~D~( \omega_{(n,l+1)}^{aligned}, \Omega_{l+1} ) \rightarrow 0. \\
		\omega_{(n, l)}^{aligned} = & \Pi_{trans}^T\omega_{(n,l)}, s.t.~D~( \omega_{(n,l)}^{aligned}, \Omega_{l} ) \rightarrow 0.
	\end{split}
	\label{wosa}
\end{equation}
	With the minimum layer-wise matrix similarity distance, the distributed weights $\omega_{(n,l,i)}$ with corresponding information are expected to be generally aligned by identifying the re-permutation matrix $\Pi_{trans}$ before each global averaging operation.
	This can be translated to an optimization problem and be resolved by different algorithms like Bipartite matching, Wassertein barycenter and Hungarian matching~\cite{opt_trans, fedma}.

\subsection{Limitations of WLA}
	Although current WLA works' significant performance escalation demonstrated the necessity of parameter alignment for FL, there is an essential question:
	\textbf{\textit{Does the weight matrix distance really reflect the information mismatching across distributed DNN models?}}
%
Current methods' alignment accurateness highly depends on the selected similarity metrics and their operation targets, \textit{e.g.}, \textit{MSE} and Euclidean distance on weight or activation matrices~\cite{fedma, rmatching}.
	However, these quantitative alignment criteria may not fully match the weights carrying the same learning feature information.
	Many recent works have demonstrated the necessity of qualitative parameter interpretation and feature visualization for DNN design and optimization~\cite{bmvc,vis1,vis3}.
%
Furthermore, the practical FL may involve non-IID local data and even non-IID learning classes across nodes.
	In such cases, parameters will encode non-uniformed information and can be not fully-matched at all~\cite{non-iid0}, and a forced value-based matching can cause catastrophic performance degradation.

Besides the alignment criteria, we would like to ask another question:
	\textit{\textbf{How to effectively and efficiently guarantee the alignment across the federated learning process?}}
Most prior works adopt a post-alignment method~\cite{bayesian, pfnm, fedma, rmatching}, which analyzes and match parameters across models before every global averaging operation, resulting in heavy computation workloads.
	And the parameter similarity analysis also requires activation data sharing that can compromise the input data privacy.

\section{Feature-Level Alignment (FLA): \\ A New Perspective}

We expect to address the above problems through a series of technical contributions:
	We first answer the ``\textit{what-to-align}'' question by promoting the previous weight-level similarity-based parameter alignment to a feature level;
	We then answer the ``\textit{how-to-align}'' question by proposing a feasible feature allocation scheme to establish firm correlations between DNN structures and their designated assigned learning features;
	Eventually, we design a feature-aligned FL framework --- \systemname, which enables accurate feature alignment and thus achieves superior performance than FedAvg as well as other WLA methods.
	Specifically, in this section, we interpret the feature information learned by DNN parameters and propose a novel feature-aligned learning objective for FL frameworks.

\vspace{-2mm}
\subsection{Feature Definition and Interpretation}
	Many prior works have elaborated neural networks' feature information from many perspectives\footnote{Here we consider image classification as our major deep learning task.}.
	For example, \cite{am} utilizes activation maximization to visualize each neuron's preferred input pattern, \cite{vis4} introduces the activation-based attention mechanism to illustrate neuron's region of interest in the input, and \cite{bmvc,vis1,vis2,vis4} uses the learning class preference to illustrate the neuron functionality.
	Unlike conventional quantitative approaches, these feature-oriented analysis methods provide qualitative and explicit interpretations of the DNN learning process's intrinsic mechanism.

\begin{figure}
	\centering
	\includegraphics[width=0.95\linewidth]{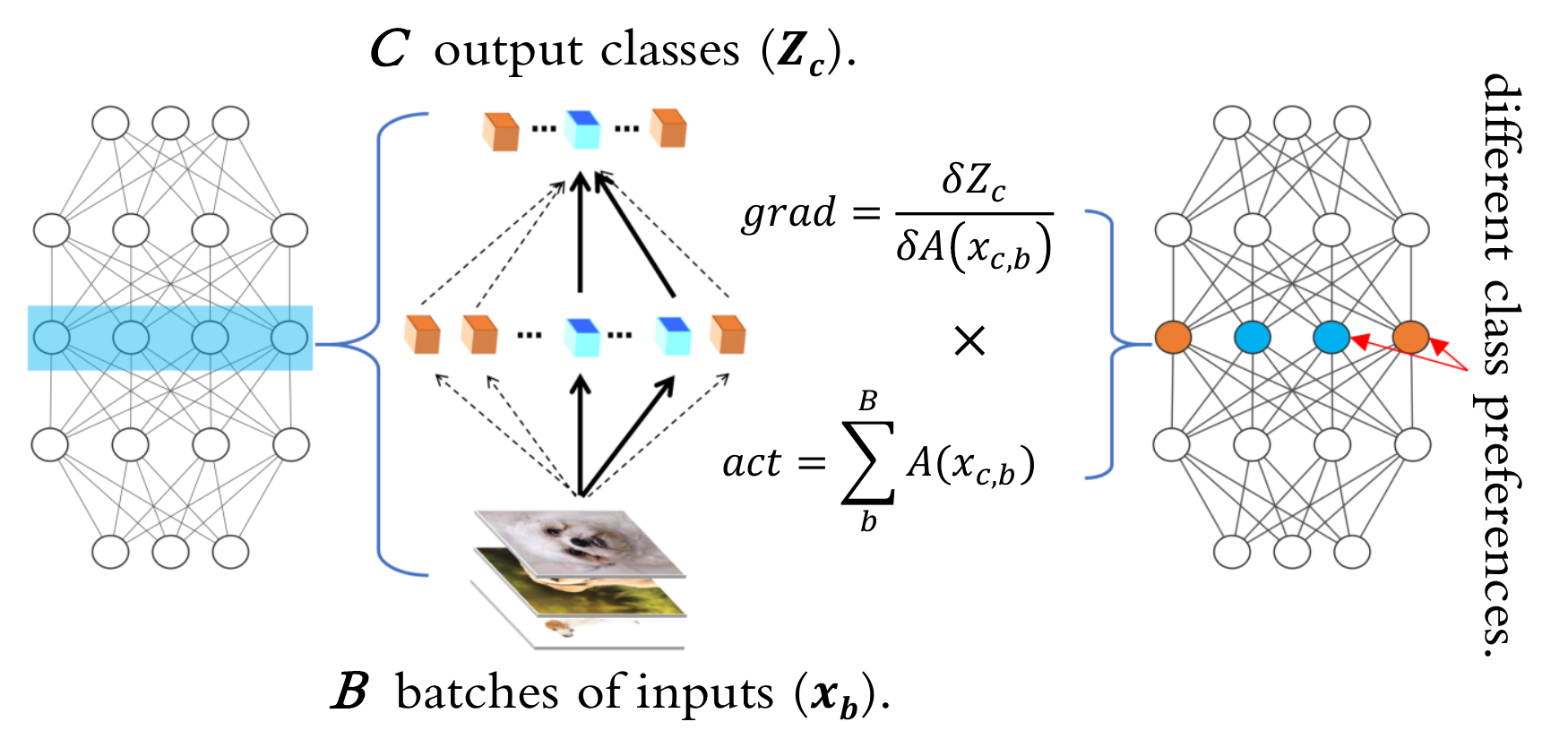}
	\vspace{-3mm}
	\caption{Activation and Gradient based Feature Analysis.}
	\vspace{-3mm}
	\label{fig:act_grad}
\end{figure}

In this work, we adopt neurons as the basic feature learning units\footnote{Here the \textit{neuron} is defined one convolutional filter if in the convolutional layer, or one neuron if in the fully connected layer.}, and practice the feature interpretation as follows:
	As shown in Fig.~\ref{fig:act_grad}, one individual neuron's learning preference can be measured by observing the neuron's activation response $A(x_c)$ on inputs $x$ from different $C$ classes, as well as its gradients {\small $\partial Z_c / \partial A(x_c)$} towards a class $c$'s prediction confidence $Z_c$.
	Combining these two factors and further generalizing to a multi-layer convolutional neural network, a neuron's learned feature information can be formulated as \textit{\textbf{a class preference vector}}:
\begin{equation}
\small
	P = [p_1,\ldots, p_c,\ldots,p_C], \hspace{2mm}\text{where}~p_{c} = \sum_b^B A(x_{c,b}) * \frac{\partial Z_c}{\partial A(x_{c,b})},
	\label{eq:feature}
	\vspace{-1mm}
\end{equation}
\normalsize
where $A~(x_{c,b})$ denotes activations and {\small $\partial Z_c/\partial A(x_{c,b})$} denotes gradients from class $c$'s confidence, both of which are averaged on $B$ batch trials.
	For each neuron, the largest index $Argmax_i(P_i)$ of feature vector $P$ indicates its primary learning class target.
	Assembling all neurons' top preferred classes together, a layer's feature encoding vector could be then obtained.

\begin{figure}
	\vspace{3mm}
	\centering
	\includegraphics[width=1\linewidth]{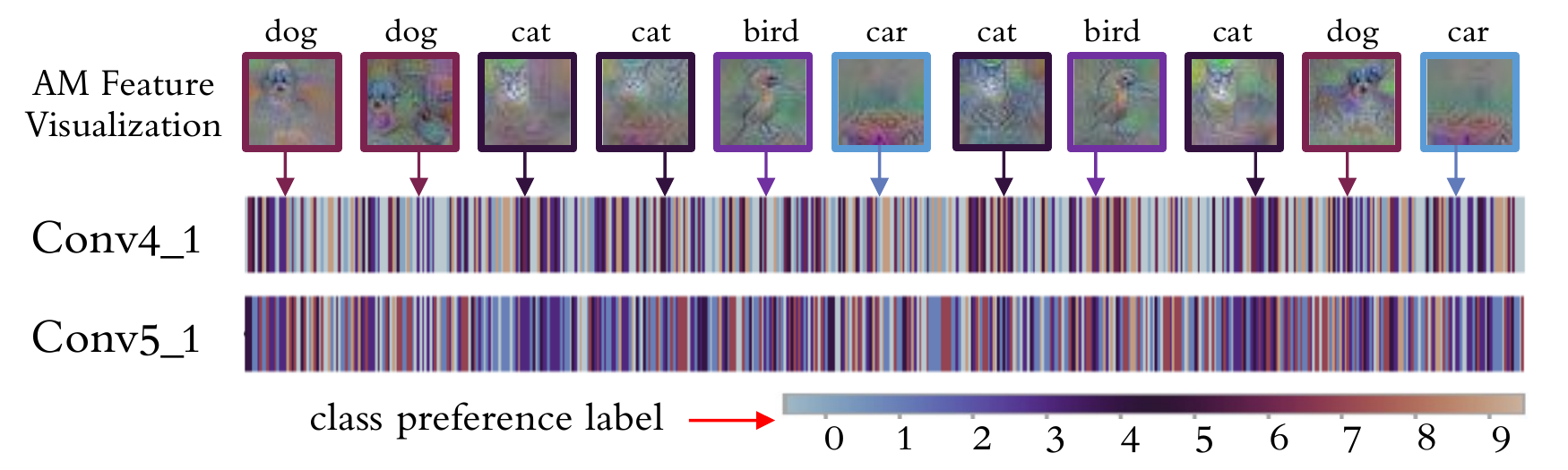}
	\vspace{-7mm}
	\caption{Class Preference Vector Visualization (VGG16 on CIFAR10).}
	\vspace{-3mm}
	\label{fig:act_grad_example}
\end{figure}

	As an example, we visualize two convolutional layers' learned feature information in Fig.~\ref{fig:act_grad_example}.
	Similar to Fig.~1, each neuron is represented by one vertical color bar, while the color denotes different primary preferred classes.
	In practice, we find such a feature interpretation method aligns well with previous AM visualization~\cite{am}, which demonstrates the effectiveness of our feature interpretation.


\subsection{Feature-Aligned FL Objective}
Based on such a feature interpretation perspective, we can re-examine the model fusion process, starting with the coordinate-based FedAvg method:
	Across the \textit{y}-dimension of Fig.~\ref{fig:motivation} (a), we sample and visualize the neuron learning class preferences from the same layer's feature encoding across all ten local DNN models.
	As we could see, neurons with the same coordinates show dramatically different class preferences.
	Since FedAvg adopts a coordinate-based averaging, massive feature averaging conflicts can happen.
	The encoded information from partial nodes can thus be lost, leading to a slower convergence rate or lower accuracy.

To alleviate the feature fusion conflicts, we propose to conduct feature-level alignment during the FL model averaging process.
Formally, the \textit{\textbf{feature-aligned federated learning objective}} could be defined as minimizing the total feature-level parameter variance among $N$ collaborative nodes :
\begin{equation}
	Minimize ~~\sum D~(p_{(n_1,l,i)}, ~~p_{(n_2,l,i)}), ~~\forall n_1, n_2 \in (1, ~N), ~~n_1 \neq n_2,
	\label{eq:objective}
\end{equation}
where $p_{(n,l,i)}$ is the feature learned by the $i^{th}$ neuron of the $l$-th layer from the $n^{th}$ node, and $D$ is an appropriate distance metric for feature vector similarity evaluation.

Naively, to solve Eq.~\ref{eq:objective}, we could calculate the feature vector of each neuron $p_{(n,l,i)}$ and then conduct post-alignment by neuron re-permutation just like previous weight-matching methods.
	However, such an approach suffers from the same limitations like inaccurate feature similarity metrics and heavy post-matching computation overheads, \textit{etc}.
	Therefore, we propose a set of novel feature alignment method by establishing firm correlations between DNN structures and their designated assigned learning features without tedium feature allocation and analysis effort.

\section{Structural Feature Allocation}
\label{sec:4}

In this section, we then answer the \textit{``how-to-align''} question, \textit{i.e.}, to design effective ways for feature-level alignment.
	Specifically, we propose a ``structural feature allocation'' scheme to establish firm correlations between DNN structures and their designated learning features:
	Given a DNN model, we adapt the model structure by constructing separated parameter groups with the group convolution method;
	Different learning classes are then assigned into individual parameter groups through a gradient redirection method;
	Therefore, we enforce one parameter group to learn a set of designated classes, thus achieving the structural feature allocation at the early stage of model training.
	When deployed into an FL scenario, such a structural feature allocation can guide explicit feature-level alignment with parameter structure group matching and provide the methodology foundation for later feature-aligned federated model fusion.

\subsection{Feature Isolation by Group Convolution}
How to guide particular features to be learned by designated parameters is the key to structural feature allocation.
	And the primary motivation of such an approach is that we notice the group convolution structure could isolate the feature distributions with separated activation forwarding and gradient backwarding processes~\cite{alexnet,shufflenet}.

\begin{figure}
\centering
\includegraphics[width=1\linewidth]{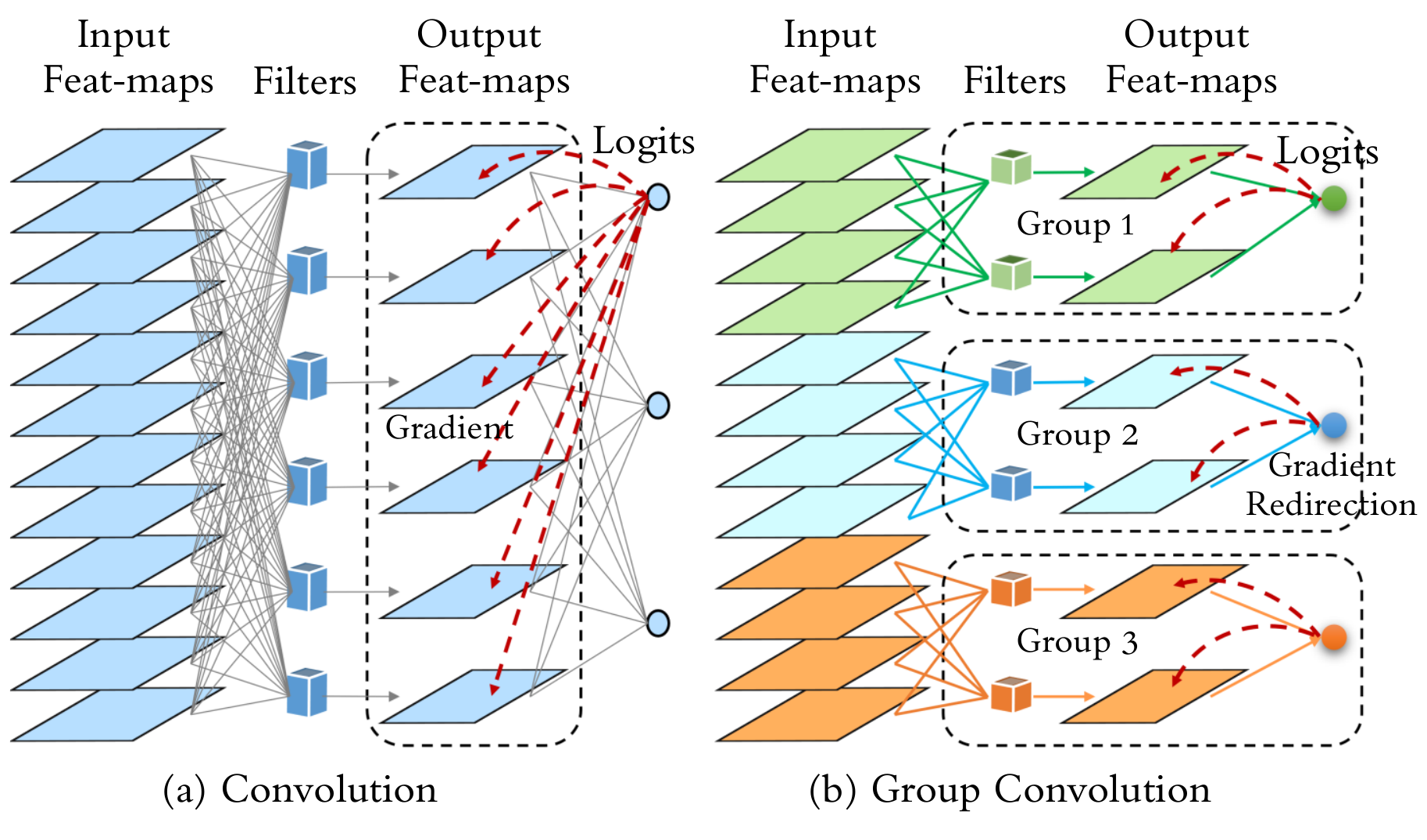}
\vspace{-7mm}
\caption{We achieve structural feature allocation by adopting group convolution and decoupled logit (fully-connected) layers. Specifically, we use group convolution for feature isolation and combine it with decoupled logit layers for feature allocation.}
\vspace{-4mm}
\label{fig:group_conv}
\end{figure}

%
Fig.~\ref{fig:group_conv} illustrates the model structure difference between common convolution and group convolution. 
	Regular convolution (Fig.~\ref{fig:group_conv} (a)) follows a densely-connected computational graph.
	By contrast, the group convolution structure (Fig.~\ref{fig:group_conv} (b)) separates the convolution operations into groups.
	The input/output feature maps are therefore mapped only within their current group.
The group convolution is first proposed in AlexNet~\cite{alexnet} by using two groups to relieve the computational burden of single GPU.
 	But after training, the model learns distinct features in two convolution groups (\textit{i.e.}, shape-oriented and color-oriented features)~\cite{alexnet}.
	Similar phenomenon is observed in ShuffleNet~\cite{shufflenet} that features become biased within each different convolutional group without shuffling.
Although initially designed for computational benefits, the grouped structure naturally demonstrates feature regulation and isolation effects in both models, which have been rarely explored in priori works.

Our hypothesis of such feature isolation phenomenon is due to the gradient isolation effect incurred by the separable computational graph in group convolution. 
As different groups are forwarding separately, the backward gradients carrying feature information also flows only within their own groups, thus gradually leading to feature isolation. 
Formally, in the regular densely-connected convolution, each output feature map $OF_i~ (i \in [1:d_o])$ is calculated by convolving on all input feature maps $IF_j~ (j \in [1:d_i])$:
\begin{equation}
OF_{1:n} = \{w_1 * IF_{1:m}, ~~w_2 * IF_{1:m}, ~~..., ~~w_n * IF_{1:m}\},
\end{equation}
where $d_o$ and $d_i$ are the output/input feature map depth, $w_i$ is weights for the $i^{th}$ convolution filter, and $*$ is the convolution operation. 
The gradients of input feature $IF_i$ can be formulated as:
\begin{equation}
\nabla IF_{j} = \sum\nolimits_i \frac{\delta OF_i} {\delta IF_j}, ~~i \in (1, ~d_o).
\end{equation}
That is, the gradient of $IF_{J}$ fuses the information from all output features ($OF_i$).
Due to the interleaved and fused gradients, the input layer's feature encoding can be highly non-predictable.
In distributed FL, such random encoding thus incurs feature mismatches and averaging conflicts, lead to sub-optimal convergence.

By contrast, the group convolution separates the computational graph, as well as the convolutional inputs \& outputs into different groups $G$. The grouped output feature maps ($OF$) are:
\begin{equation}
\begin{split}
& OF_{1:G_1} =  \{w_1 * IF_{1~:~G_1}, ~~..., ~~w_{G_1} * IF_{1~:~G_1} \} ~... \\
& OF_{G_i:G_{i+1}} = \{w_{G_i} * IF_{G_i~:~G_{i+1}}, ~~..., ~~w_{G_{i+1}} * IF_{G_i~:~G_{i+1}} \} ~...\\
& OF_{G_{g-1}:n} = \{w_{G_{g-1}} * IF_{G_{g-1}~:~m}, ~~..., ~~w_{n} * IF_{G_{g-1}~:~m}\}
			\}.
\end{split}
\end{equation}
As each input feature map contributes to the within-group output feature maps only, \textit{i.e.}, $OF_{G_i:G_{i+1}}$, the backward gradient for $IF_j$ will only fuse the information within the current group $G_{g_j}$.
\begin{equation}
\nabla IF_{j} = \sum\nolimits_i \frac{\delta OF_i} {\delta IF_j}, ~~i \in (G_{g_j}, ~G_{g_j+1}).
\end{equation}
In this case, the group convolution structure builds implicit boundaries between groups, and achieves the feature isolation effect.

\begin{figure*}[t]
	\centering
	\includegraphics[width=6.5in]{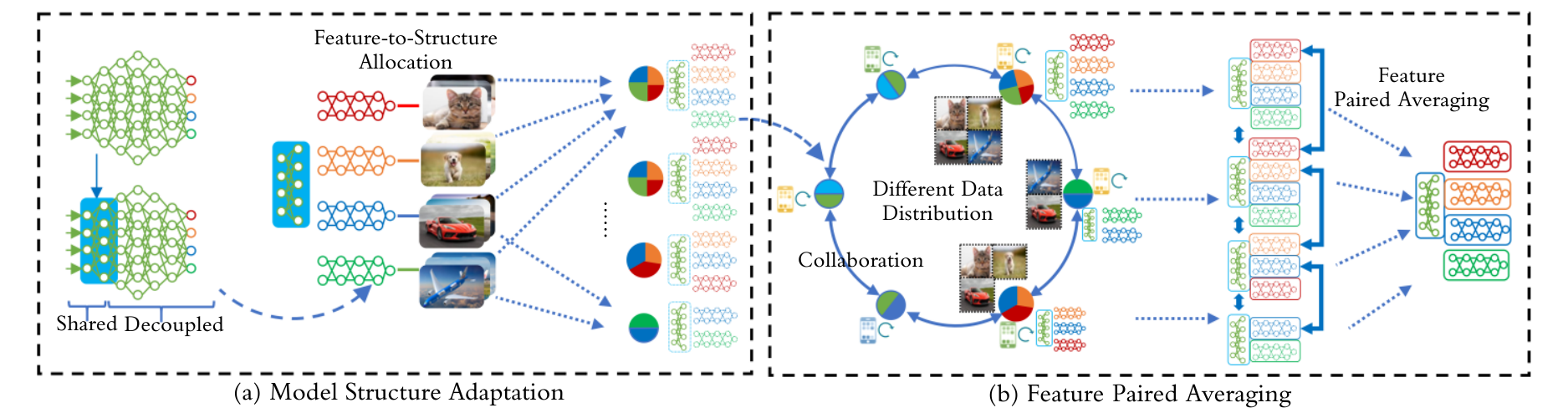}
	\vspace{-3mm}
	\caption{Our proposed Fed$^2$ framework includes two major steps: \textbf{(i)} We utilize group-convolution based structure to conduct feature-to-structure allocation; \textbf{(ii)} We then propose a feature paired averaging policy to enforce the feature alignment during federated model averaging.}
	\label{fig:system}
	\vspace{-3mm}
\end{figure*}

\subsection{Feature Allocation by Gradient Redirection}
Building upon such a feature isolation effect, we then propose a gradient redirection method to control the feature allocated in each convolution group.
	Our main idea is first separating the gradient components carrying different classes' features, and then redirecting them into different groups.
	Specifically, this is done by the decoupled fully-connected layer, in which different class logits are connected only to their corresponding group.
	Combining both feature isolation and allocation, we are able to achieve structural feature allocation and facilitate our feature-aligned federated learning framework.

As shown in Fig.~\ref{fig:group_conv} (a), traditional logit layers (\textit{i.e.}, the last fully-connected layer) usually fully connect all input feature maps (convolutional filters) with the logits.
	The gradients carrying features from each logit thus flow through all output feature maps ($OF_i$):
\begin{equation}
OF_{1...n} \leftarrow \nabla Logit_c, ~~c \in (1, ~C).
\end{equation}
$C$ is the number of logits. Due to the fully-connected computational graph, the feature encoding in each layer becomes non-predictable.

Different from that, we decouple the original logit layer into groups as well.
	An example is shown in Fig.~\ref{fig:group_conv} (b).
	Each sub-layer maps the class logit(s) to one corresponding convolutional group only, enforcing the gradients flowing backwards to the mapped structure group without any leakage:
\begin{equation}
OF_{G_i~:~G_{i+1}} \leftarrow \nabla Logit_g, ~~\text{$g$ is a subset of $(1,~C)$.}
\label{eq:assign}
\end{equation}
Here $OF_{G_i~:~G_{i+1}}$ is one group of output feature maps, and $Logit_g$ is the group of class logits that are assigned to this structure group.

By the gradient redirection, each structure group acts as the anchor for the allocated features.
During the FL training process, the feature of these classes will be continuously contained within the group, thus enforcing the structural feature allocation.

\section{Fed2 Framework}

Based upon the structural feature allocation methodology, we then propose \systemname, a feature-aligned federated learning framework, which further enhanced the feature-level alignment with a particular DNN structure design and model fusion scheme for FL. 

\subsection{FLA-Enhanced DNN Structure Design}

Fig.~\ref{fig:system} (a) shows the overview of our model structure adaptation.
	For common DNN models like VGG~\cite{vgg} and MobileNets~\cite{mobilenet}, our structural adaptation splits the model into two parts:
	For the lower convolutional layers, we maintain the densely-connected structures as shared layers.
	For the higher convolutional and fully connected layers, we transform them into the group-wise structure.

\vspace{1mm}
\noindent \textbf{Shared Layers for Feature Sharing}: 
The design of lower convolutional layers to be shared is due to the shallow layers learn mostly basic shared features, which shows few feature averaging conflicts~\cite{bmvc, vis1, vis3}. 
	In such cases, blindly separating these layers into groups can prevent low-level neurons receiving gradients from all groups, leading to bad learning performance.
	Our empirical study also verifies such conclusion (as we will show later). 
	Therefore, we reserve shallow layers as the shared layers.

\vspace{1mm}
\noindent \textbf{Decoupled Layers for Feature Isolation}: 
By contrast, for the deeper convolutional layers, the encoded features diverges more and are easier to conflict with each other during averaging~\cite{critical_path, rmatching}.
	Therefore, we adopt group convolution and construct separable structure groups in these layers for further feature alignment.

To determine an appropriate number of decoupled layers, we evaluate the feature divergence of layer $l$ by the total variance ($TV$) of all neurons' feature vectors $P_{l,i}$ (defined in Eq.~\ref{eq:feature}) in this layer:
\begin{equation}
\begin{split}
	TV_l = \sum\nolimits_i^I \frac{1}{I}|| P_{l,i} - E(P_{l,i})||_{2}.
	\label{eq:share_depth}
\end{split}
\end{equation}
$I$ is the number of neurons in layer $l$.
Such layer-wise feature total variance usually maintains low in the lower layers and surge high in later layers. 
	We therefore determine an appropriate decoupling depth by thresholding the $TV$ for the group-wise transformation.

\vspace{1mm}
\noindent \textbf{Feature-to-Structure Allocation Enhancement}: 
After determining the decoupled layers, we construct convolution groups and conducts gradient redirection by logit(s) allocation in Eq.~\ref{eq:assign}.
Such a step accomplishes an explicit feature-to-structure allocation.
One illustrative example is shown in Fig.~\ref{fig:system} (a)\footnote{For simplicity, Fig.~\ref{fig:system} (a) shows a one-class to one-group mapping example. For large datasets with more classes (\textit{e.g.}, 100 classes), multi-classes to one-group is achievable and also yields similar feature alignment benefits, as we will evaluate later.}. 
For future feature alignment, we thus can easily match different convolution group structures by the learning tasks (\textit{i.e.}, logits) mapped to them.

Another optimization is we replace the batch normalization (BN) to be group normalization (GN)~\cite{group_norm}.
Previous works have shown that BN can influence the distributed training performance as different local models tend to collect non-consistent batch mean and variance (especially in non-IID cases)~\cite{iclr}. 
Our structure design, by enforcing feature allocation, alleviates the feature statistics divergence within each group.
	Therefore, we incorporate the GN layer and further improve the model convergence performance. 
	We will demonstrate the effectiveness of GN layers in later experiments.

By the proposed structure adaptation, \systemname enables structure-feature alignment before the training process.
	Such structure-feature pre-alignment also greatly simplifies the following matching process, which alleviates the heavy distance-based optimization computation and achieves better feature alignment effect.

\subsection{Feature-Aligned Federated Model Fusion}

With the feature-to-structure pre-alignment, \systemname can then promote previous weight-level matching methods to the feature-level.
Specifically, we propose the feature paired averaging algorithm.


Firstly, the shared layers will be averaged among $N$ collaborative nodes.
	As they extract fundamental shared features with less feature conflicts, the coordinate-based FedAvg can be directly applied:
\begin{equation}
	\Omega_{shared} = E(\omega_{shared}^{n}), ~~n \in (1, ~N),
	\label{eq:sharelayer}
\end{equation}
where $\omega_{shared}^n$ denotes the weights of the shared layers from the $n$-{th} local model, and $\Omega_{shared}$ is the averaged global model weight.

For decoupled layers, as different groups are assigned with different class logits, weight averaging should be conducted within the groups that share the same learning tasks.
That is, only the groups that have the paired learning class are averaged together:
\begin{equation}
\begin{split}
	\Omega^{g} = E(\omega^g_{i,j}), ~~\text{iff} ~~Logits(\omega^g_i) = Logits(\omega^g_j), \\
	~~\forall i, j \in (1, N), ~~i \neq j.
	\label{eq:pair_avg}
\end{split}
\end{equation}
Here $\Omega^{g}$ denotes the global weights of the $g$-th group structure, and $\omega^{g}_{i,j}$ are the local model weights of group $g$ on nodes $i$ and $j$.
The $g^{th}$ group's weights from two local models will be averaged if and only if they are paired, \textit{i.e.}, $Logits(\omega_g^i) = Logits(\omega_g^j)$.

The proposed feature paired averaging method accomplishes the last step for feature aligned averaging in \systemname.
	Benefited from the explicit feature-to-structure pre-alignment, our group pairing process only needs to match the learning logits (an one-hot class vector). 
	This greatly simplifies the matching complexity than previous parameter matching like weights and activations~\cite{fedma, rmatching, pfnm}.
	Therefore, \systemname also alleviates the heavy computation and communication overhead of traditional post-alignment methods.

\section{Experiment}

We evaluate \systemname with image classification tasks on CIFAR10 and CIFAR100. 
Three DNN\ models (VGG9~\cite{fedma}, VGG16~\cite{vgg}, and MobileNetv1~\cite{mobilenet}) are adopted to evaluate the generality of our structure adaptation method.
Without specific mentioning, all baselines use the original network, while \systemname adopts a general decoupling step, \textit{i.e.}, decoupling the last 6 layers with 10 convolution groups for three networks.
%
For local data distributions, we consider both IID and non-IID scenarios.
%
State-of-the-art (SOTA) works including FedAvg~\cite{fl}, FedMA~\cite{fedma} and FedProx~\cite{fedprox} are compared to demonstrate the training efficiency and convergence benefits of our framework.

\begin{figure}[!b]
	\centering
	\vspace{-4mm}
	\includegraphics[width=0.9\linewidth]{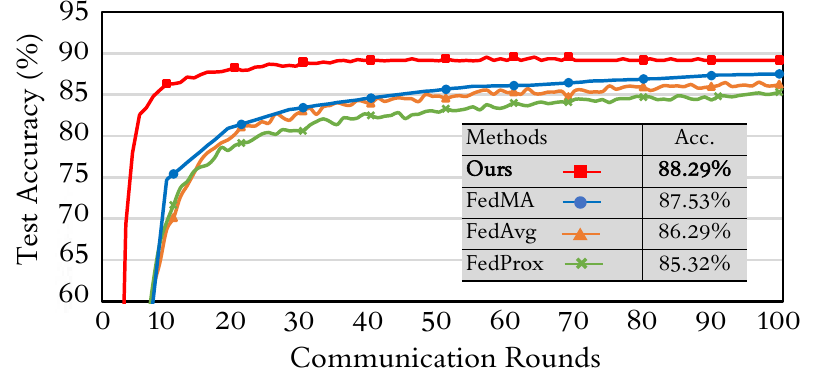}
		\vspace{-3mm}
	\caption{Communication Efficiency Comparison.}
		\vspace{-3mm}
	\label{fig:communication}
\end{figure}

\begin{figure}[!tb]
\includegraphics[width=0.9\linewidth]{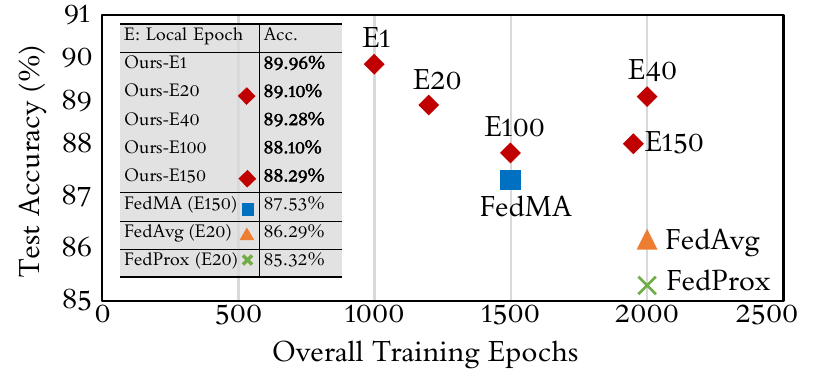}
\vspace{-3mm}
\caption{Computational Efficiency Comparison.}
\label{fig:computation}
\end{figure}

\subsection{FL Convergence Performance}

We first compare the convergence performance of \systemname with other SOTA methods.
The experimental settings are kept same with~\cite{fedma} using VGG9 on CIFAR10 dataset.
	The heterogeneous data partition with $J ~(J=16)$ clients are adopted by sampling $p_c \sim Dir_J (0.5)$ and allocating a $p_{c,j}$ proportion of the training data of class $c$ to local client $j$, where $Dir_j (0.5)$ is the dirichlet data distribution.

We evaluate the FL convergence performance from two aspects: 
	(1) convergence rate: accuracy w.r.t. communication rounds; and
	(2) computation efficiency: accuracy w.r.t. computational efforts.

\vspace{1mm}
\noindent \textbf{Convergence Rate.}
Fig.~\ref{fig:communication} compares the test accuracy curves w.r.t communication rounds between \systemname (red line) and other methods.
	As we can see, our \systemname shows superior convergence rate compared to the other three methods. 
	With roughly 40 rounds, our method achieves the best accuracy 88.29\%.
	In contrast, other methods can take 100 rounds but still achieve lower accuracy, \textit{e.g.}, FedMA (87.53\%, -0.76\% than ours) and FedAvg (86.29\%, -2.0\% than ours).

\vspace{1mm}
\noindent \textbf{Computation Efficiency.}
We further demonstrate the computation efficiency of \systemname by comparing the model accuracy w.r.t the overall computational workloads.
Here the computational workloads are measured by the overall local training epochs on all nodes.
The results are shown in Fig.~\ref{fig:computation}.
As other methods' accuracies reported in ~\cite{fedma} are with varied local epoch settings ($E$), we conduct \systemname in different settings for fair comparison.

\systemname achieves better accuracy (89.1\%) than FedAvg and FedProx under E=20 settings, and meanwhile use less computation efforts (1200 vs. 2000 epochs).
Compared to FedMA under E=150 setting, \systemname finally achieves 88.29\% accuracy, {+0.76\%} better than FedMA (87.53\%) with slightly higher training efforts (2000 vs. 1500 epochs).
Furthermore, \systemname's optimal model accuracy achieves {89.96\%} at E=1 setting, which surpasses all other methods' accuracy by large margins (\textit{e.g.}, {+2.32\%} than FedMA) and meanwhile consumes the least training workloads (1000 epochs).

\begin{figure*}[]
\centering
\includegraphics[width=6.4in]{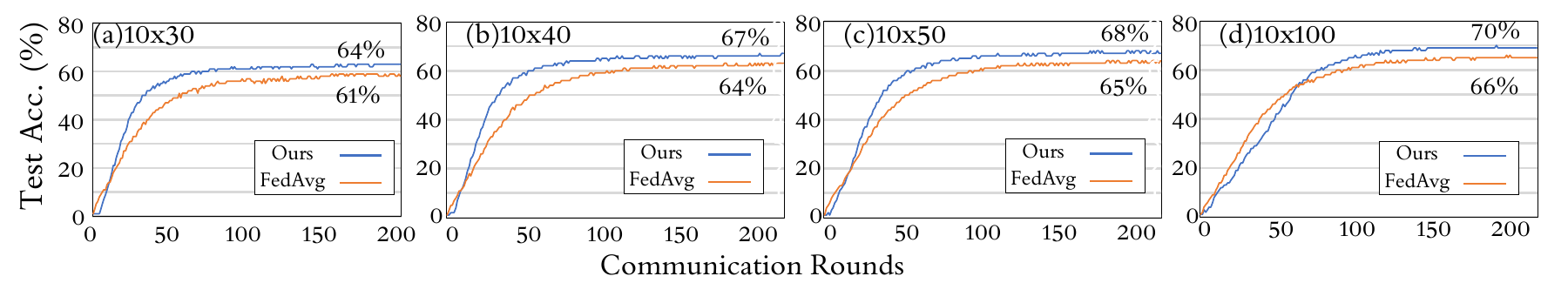}
\vspace{-2mm}
\caption{Convergence Speed Comparison between FedAvg and our proposed framework ({VGG16 on CIFAR100}).}
\label{fig:conv_speed}
\vspace{-3mm}
\end{figure*}

\begin{table}[]
\caption{Data Heterogeneity (N: \# of nodes. C: \# of classes).}
\vspace{-2mm}
\renewcommand\arraystretch{0.9}
\setlength{\tabcolsep}{1.5mm}{
\begin{tabular}{cccccc}
\toprule
   CIFAR10               & N * C & 10$\times$3 & 10$\times$4 & 10$\times$5 & 10$\times$10 \\ \midrule
\multirow{2}{*}{VGG9}      & FedAvg~\cite{fl}         & 82\% & 84\% & 85\% & 88\%  \\ 
                           & \textbf{Ours}           & 83\% & 88\% & 88\% & 90\%  \\ \midrule
\multirow{2}{*}{MbNet}   & FedAvg~\cite{fl}         & 67\% & 71\% & 79\% & 85\%  \\ 
                           & \textbf{Ours}           & 86\% & 88\% & 90\% & 91\%  \\ \bottomrule \toprule
   CIFAR100              & N * C & 10$\times$30 & 10$\times$40 & 10$\times$50 & 10$\times$100 \\ \midrule
\multirow{2}{*}{VGG16}      & FedAvg~\cite{fl}   & 61\% & 64\% & 65\% & 66\%  \\ 
                           & \textbf{Ours}           & 64\% & 67\% & 68\% & 70\%  \\ 
\bottomrule
\end{tabular}
\label{table:data_heter}}
	\vspace{-6mm}
\end{table}

\vspace{-2mm}
\subsection{Scalability Evaluation}

We then conduct scalability evaluation to demonstrate the generality of \systemname in varied experimental settings.
Specifically, we consider four scalability dimensions:
(i) data heterogeneity scaling from IID to Non-IID;
(ii) learning task complexity with different number of leaning classes;
(iii) FL system complexity with different number of nodes; and 
(iv) low to high FL communication frequencies.

\vspace{1mm}
\noindent \textbf{Data Heterogeneity (IID to Non-IID).}
We first show that \systemname provides consistent accuracy improvement under full-spectrum data heterogeneity in Table~\ref{table:data_heter}.
	The experimental setting $N * C$ indicates there are $N$ nodes, and each node has only $C$ classes present in the local data.
	A smaller $C$ means the data distribution on local nodes are more skewed, which usually leads to lower accuracy in FL.

Table~\ref{table:data_heter} shows the FL performance of VGG9 and MobileNet on CIFAR10. 
	Our \systemname framework consistently outperforms FedAvg by large margins.
	Specifically on VGG9, \systemname achieves \textbf{+1\%$\sim$+4\%} accuracy improvement across all heterogeneity settings. 
Meanwhile, we notice that MobileNet suffers more from the highly-skewed non-IID data. 
	Under the $10\times3$ setting, FedAvg on MobileNet only achieves 67\% accuracy. 
	By contrast, \systemname achieve 86\% accuracy, \textbf{+19\%} than FedAvg (67\%).
	The underlying reasons of the accuracy improvement is due to the structurally aligned feature distribution across different local models, as demonstrated in Fig.~\ref{fig:motivation} (c).
	Such feature alignment alleviates the feature-level averaging conflicts and thus provides models higher convergence accuracy.

\vspace{1mm}
\noindent \textbf{Classification Complexity.}
We then evaluate \systemname using VGG16 on CIFAR100 with more classification classes. 
	Full-spectrum data heterogeneity settings from $10\times30$ to $10\times100$ are used.
	As we can see from Table~\ref{table:data_heter}, similar conclusion could be drawn that \systemname consistently outperforms FedAvg by \textbf{+3\%$\sim$+4\%} accuracy.

Besides that, Fig.~\ref{fig:conv_speed} shows the test accuracy curves in the training process for both methods.
	In all non-IID settings (a-c), \systemname consistently shows higher convergence speed using only 50-80 rounds to achieve convergence, while FedAvg usually needs at least 100 rounds. 
	One exception is the 10x100 IID setting (d), the initial convergence rate of FedAvg is slightly faster, potentially because the IID data distribution leads to less feature divergence in the beginning stage of FL. 
	Nevertheless, our method soon exceeds FedAvg after 50 epochs and finally achieves \textbf{+4\%} accuracy than FedAvg, showing the necessity of feature alignment in achieving the optimal model convergence accuracy.

\begin{table}
	\caption{Node Scalability (N: \# of nodes. C: \# of classes).}
	\vspace{-2mm}
	\renewcommand\arraystretch{0.9}
	\setlength{\tabcolsep}{2.5mm}{
	\begin{tabular}{cccccc}
		\toprule
		& N * C & 10$\times$5 & 20$\times$5 & 50$\times$5 & 100$\times$5 \\ \midrule
		\multirow{2}{*}{VGG9}      & FedAvg~\cite{fl}         & 85\% & 86\% & 83\% & 83\%  \\
		& \textbf{Ours}           & 88\% & 88\% & 86\% & 87\%  \\ \midrule
		\multirow{2}{*}{MbNet} 	   & FedAvg~\cite{fl}         & 79\% & 85\% & 81\% & 78\%   \\
		& \textbf{Ours}           & 90\% & 90\% & 89\% & 88\%   \\ \bottomrule
	\end{tabular}}
	\vspace{-5mm}
	\label{table:nodes}
\end{table}

\vspace{1mm}
\noindent \textbf{Node Scalability.}
We then evaluate the scalability of \systemname with the increasing number of FL nodes.
	Specifically, we scale up the number of collaborative nodes from 10 to 100 with one medium data heterogeneity setting (each node only have 5 classes in the local data distribution).
The results are shown in Table~\ref{table:nodes}. 
	Without loss of generality, \systemname provides consistently better performance ranging from \textbf{+2\%$\sim$4\%} on VGG9 and \textbf{+5\%$\sim$11\%} on MobileNetV1.

\vspace{1mm}
\noindent \textbf{Communication Frequency.}
We finally evaluate the performance of \systemname under different communication frequencies.
	Here we use communication per epochs ($E$) to indicate the frequency.
	A larger $E$ indicates a lower frequency. 
	In such cases, FL performance usually becomes worse since the model collaboration are less frequent, which can incur severer feature divergence.

\begin{figure}[!tb]
\includegraphics[width=1\linewidth]{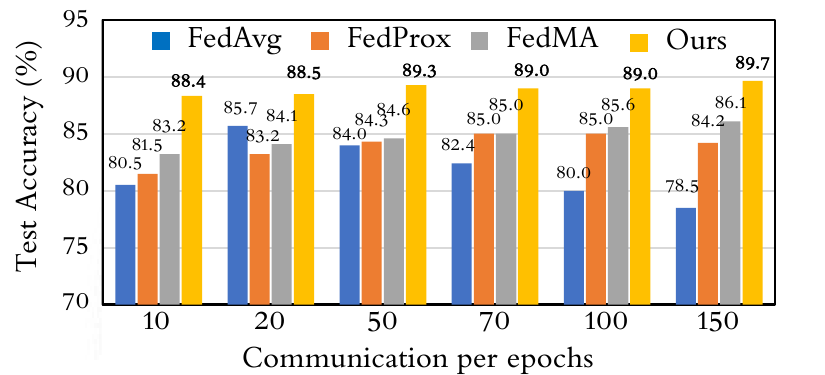}
\vspace{-6mm}
\caption{Communication Frequency Comparison.}
\label{fig:frequency}
\vspace{-5mm}
\end{figure}

Fig.~\ref{fig:frequency} compares \systemname with other methods under different communication frequencies.
	All models are trained with 54 rounds as per settings in~\cite{fedma}.
	As we can see, FedAvg (blue bar) shows lower accuracy (85.7\%$\rightarrow$78.5\%) when the frequency decreases from once per 20 epochs to once per 100 epochs.
	In contrast, \systemname with feature alignment averaging is not affected by the lower communication frequency, showing continuously the best performance (\textbf{88\%$\sim$90\%}) and improving FedMA by \textbf{+3.4\%$\sim$5.1\%} accuracy under all settings.

\subsection{Sensitivity Analysis}

We finally conduct sensitivity analysis on three design components in \systemname, including different sharing layer depth, different number of groups, and the group normalization optimization.

\begin{figure}[!tb]
	\centering
	\includegraphics[width=0.85\linewidth]{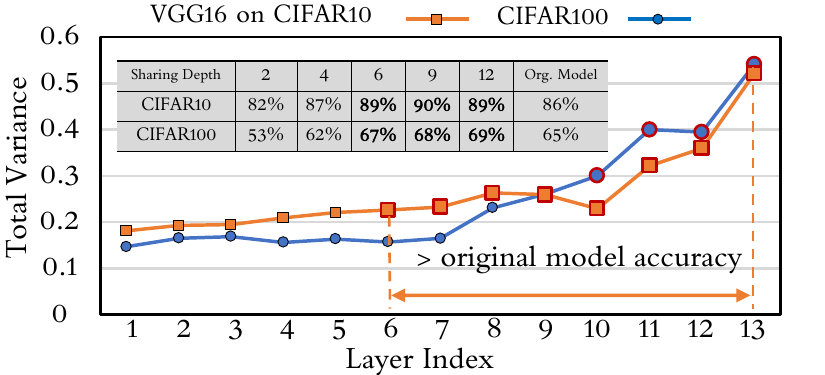}
	\vspace{-3mm}
	\caption{Sharing Depth Analysis.}
	\label{fig:depth}
	\vspace{-5mm}
\end{figure}

\vspace{1mm}
\noindent \textbf{Sharing Depth Analysis.}
We first demonstrate that our framework's performance is robust to the sharing depth hyper-parameter selection in Fig.~\ref{fig:depth}.
	As we could observe, the total variance of the layers (from a pre-trained model with 50 epoch pre-training) offers good indication of the layer-wise feature divergence.
	The results also show that it is necessary to keep enough layers (4$\sim$6) shared so that the fundamental features could be better learned by all nodes' collaboration. 
	By retaining enough shared layers in our design, \systemname's performance is highly robust to the sharing-depth hyper-parameter selection, achieving consistently better accuracy than original non-grouped model in a wide range (\textit{e.g.}, 6 $\sim$ 13).

\begin{figure}[!tb]
	\centering
	\includegraphics[width=0.8\linewidth]{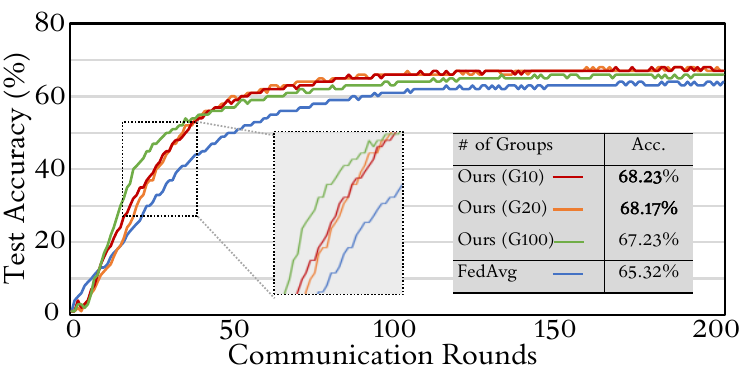}
	\vspace{-3mm}
	\caption{Number of Groups Analysis.}
		\vspace{-5mm}
	\label{fig:groups}
\end{figure}

\vspace{1mm}
\noindent \textbf{Grouping Number Analysis.}
Similar analysis shows \systemname's performance robustness w.r.t different number of groups selection.
	The results are shown in Fig.~\ref{fig:groups} (VGG16 on CIFAR100, N*C:10$\times$50).

We evaluate three group settings (G=10, 20, 100).
Overall, three settings all show better accuracy than FedAvg, demonstrating the effectiveness of using group convolution for feature alignment.
	Among them, G=10 and G=20 achieve the optimal accuracy at $\sim$68\%, +2.7\% than FedAvg with non-grouped structure (65.3\%).
G=100 setting, though achieving sub-optimal accuracy improvement (+1.9\% than FedAvg), shows the best convergence speed in the early stage (the green curve).
	We hypothesize this is due to its most fine-grained feature allocation and alignment effect (The G=100 settings enable one-class to one-group mapping, while others are multi-class to one-group mapping).
	However, with too many groups split, the per-group capacity (\textit{e.g.}, \# of neurons in each group) becomes limited, which slightly hinders the final convergence accuracy. 
	
\begin{figure}[!tb]
	\centering
	\includegraphics[width=0.8\linewidth]{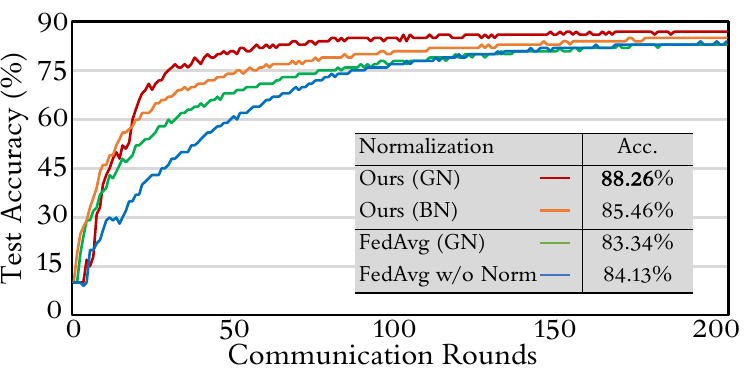}
	\vspace{-3mm}
	\caption{Normalization Strategy Analysis.}
		\vspace{-5mm}
	\label{fig:batchnorm}
\end{figure}

\vspace{1mm}
\noindent \textbf{Normalization Strategy Analysis.}
We finally conduct analysis on our normalization strategies (VGG9, CIFAR10, N*C:10x4) in Fig.~\ref{fig:batchnorm}.
	The baseline FedAvg without norm achieves 84.13\% accuracy. 
	FedAvg+GN hurts the model performance, degrading the accuracy to 83.34\%. 
	By contrast, ours+GN achieves the best accuracy 88.26\%, +2.8\% than ours+BN (85.46\%). 
	This implies that our grouped model structure indeed incurs less statistics divergence within each group, thus GN could boost the FL performance.

\section{Conclusion}
\label{sec:conc}

In this paper, we proposed \systemname, a feature-aligned federated learning framework to resolve the feature fusion conflicts problem in FedAvg and enhance the FL performance.
	Specifically, a feature interpretation method is first proposed to analyze the feature fusion conflicts.
	To alleviate that, we propose a structural feature allocation methodology by combining feature isolation and gradient redirection.
	The \systemname framework is then proposed, which composed of (i) model structure adaptation and (ii) feature paired averaging, to achieve firm feature alignment throughout the FL process.
	Experiment demonstrates significant improvement in convergence speed, accuracy and computation/communication efficiency than state-of-the-art works.

\vspace{5mm}
{
\bibliographystyle{ieee_fullname}
\bibliography{_ref/dac_shuffled}
}

\end{document}